\def\year{2021}\relax
\documentclass[letterpaper]{article} 
\usepackage{aaai21}  
\usepackage{times}  
\usepackage{helvet} 
\usepackage{courier}  
\usepackage[hyphens]{url}  
\usepackage{graphicx} 
\usepackage{natbib}  
\usepackage{caption} 
\usepackage{multirow}
\usepackage{amsmath}
\usepackage{subfigure}

\title{Prediction of Landfall Intensity, Location, and Time of a Tropical Cyclone}
\author{

    Sandeep Kumar\textsuperscript{\rm 1, 2}, Koushik Biswas \textsuperscript{\rm 1}, Ashish Kumar Pandey\textsuperscript{\rm 3}
    \\
}
\affiliations{

    \textsuperscript{\rm 1}Department of Computer Science, IIIT Delhi, India 110020\\
    \textsuperscript{\rm 2}Department of Mathematics, Shaheed Bhagat Singh College, University of Delhi, India 110017\\
    \textsuperscript{\rm 3}Department of Mathematics, IIIT Delhi, India 110020\\
    sandeep\_kumar@sbs.du.ac.in,
    koushikb@iiitd.ac.in,
    ashish.pandey@iiitd.ac.in

}

\begin{document}

\maketitle

\begin{abstract}

The prediction of the intensity, location and time of the landfall of a tropical cyclone well advance in time and with high accuracy can reduce human and material loss immensely. In this article, we develop a Long Short-Term memory based Recurrent Neural network model to predict intensity (in terms of maximum sustained surface wind speed), location (latitude and longitude), and time (in hours after the observation period) of the landfall of a tropical cyclone which originates in the North Indian ocean. The model takes as input the best track data of cyclone consisting of its location, pressure, sea surface temperature, and intensity for certain hours (from 12 to 36 hours) anytime during the course of the cyclone as a time series and then provide predictions with high accuracy. For example, using 24 hours data of a cyclone anytime during its course, the model provides state-of-the-art results by predicting landfall intensity, time, latitude, and longitude with a mean absolute error of 4.24 knots, 4.5 hours, 0.24 degree, and 0.37 degree respectively, which resulted in a distance error of 51.7 kilometers from the landfall location. We further check the efficacy of the model on three recent devastating cyclones Bulbul, Fani, and Gaja, and achieved better results than the test dataset.
\end{abstract}

\section{Introduction}

Tropical cyclones (TC) are one of the most devastating natural phenomenon that originates on tropical and subtropical waters and a commonly occurring natural disaster in coastal area. TC is characterised by warm core, and a low pressure system with a large vortex in the atmosphere.  TC brings strong winds, heavy precipitation and high tides in coastal areas and resulted in huge economic and human loss. Over the years, many destructive TCs have originated in the North Indian Ocean (NIO), consisting of the Bay of Bengal and the Arabian Sea. In 2008, Nargis, one of the disastrous TC in recent times, originated in the Bay of Bengal and resulted in 13,800 casualties alone in Myanmar and caused US\$15.4 billion  economic loss \cite{cyclonefhfc2009}. In 2018, Fani cyclone caused 89 causalities in India and Bangladesh, and US\$9.1 billion economic loss \cite{turbulanceksp2020}. Indian Meteorological department (IMD) defines the  intensity  of  a TC in terms of ``Grade”, which is derived  from the 
ranges  of  the  Maximum  Sustained Surface  Wind  Speed  (MSWS)\footnote{FAQ on Tropical Cyclones - \url{http://www.rsmcnewdelhi.imd.gov.in/images/pdf/cyclone-awareness/terminology/faq.pdf}} as shown in Table \ref{grade}. The prediction of cyclone's trajectory and intensity is crucial to save both material loss and human lives. Track and intensity prediction of a cyclone, well advance in time is not an easy task because of the complex and non-linear relationship between its cause factors which also include external factor like terrain.

\begin{table}[t]
    \centering
    \begin{tabular}{|c|c|c|} \hline
     {\bf Grade} & {\bf Low pressure system}   &  {\bf MSWS (knots) }\\ \hline
         0 & Low Pressure Area (LP) & $<$17\\
1 & Depression (D) & 17-27\\
2 & Deep Depression (DD) & 28-33\\ 
3 & Cyclonic Storm (CS) & 34-47\\ 
4 & Severe Cyclonic Storm (SCS) & 48-63\\ 
5 & Very Severe CS (VSCS) & 64-89\\ 
6 & Extremely Severe CS (ESCS) & 90-119\\
7 & Super Cyclonic Storm (SS) & $\geq$120\\  \hline
    \end{tabular}
\caption{The Grade classification of the low pressure systems by IMD.}
    \label{grade}
\end{table}

The most widely used cyclone track and intensity prediction techniques can be classified into statistical, dynamical, and ensemble models\footnote{Track and Intensity models - \url{https://www.nhc.noaa.gov/modelsummary.shtml}}. However these methods have their own limitations in terms of huge computation power required and the requirement of long duration data \cite{whaw2009statistical, htjs2007statistical, kkcdsss1999improved}. In recent years, with the increase in computational power and availability of huge data, new models using Artificial Neural Networks (ANNs) have been increasingly used to forecast track and intensity of cyclones \cite{LEROUX201885, abjg2018predicting, gsmgmc2020learning, moradi2016sparse}. 

The most important prediction about a TC is its arrival at land, known as landfall of a cyclone. The accurate prediction about the location and time of the landfall, and intensity of the cyclone at the landfall will hugely help authorities to take preventive measures and reduce material and human loss. In this work, we attempt to predict intensity, location, and time of the landfall of a TC at any instance of time during the course of a TC by observing the cyclone for as few number of hours (h) as possible. We have  build a model using Long Short-Term Memory (LSTM) based Recurrent Neural Network  (RNN) which can provide predictions about the landfall of a cyclone originating in the NIO. The developed model uses 12h, 18h, 24h or 36h data of a TC, anytime during the course of a TC and predicts the intensity, location, and time of the landfall. In subsequent sections, we will describe the related work, methodology, and data used in this study. Finally, we will compare our results with the one reported by IMD.

\section{Related Work}
In \cite{chaudhuri2015track, chaudhuri2017swarm}, authors presented an ANN to predict the intensity and track of a TC in NIO using cyclones data from 2002 to 2010. In \cite{mbn2013eval}, authors have presented the current state-of-the-art track prediction accuracy in terms of distance error between predicted location and actual location of a TC, achieved by IMD for cyclones originated in NIO. In all these works, the predictions have been provided for a certain number of lead hours say 6h, 12h, or 24h and do not specifically focus on predicting the intensity, location, and time of the landfall of a TC. Moreover, these works do not use the complete data available on the IMD website and restrict to certain number of years to obtain their dataset. In comparison, we study the prediction problem at the landfall which is more challenging as a TC may behave abruptly close to the landfall. Also, we do a more comprehensive study by including all available data on the IMD website. We compare our results directly with the predictions achieved by IMD for the landfall of a TC in recent years. In recent works \cite{gsmgmc2020learning, giffard2018deep, maskey2020deepti, RRMRD2018}  TC's track and intensity prediction problem is targeted using reanalysis and satellite data.

\section{Model}

We tried various machine learning models like ANN, RNN (based on GRU, LSTM and BiLSTM) and 1D-CNN for the above stated prediction problems. RNN model based on LSTM gives the best result. In this section, we will briefly describe the LSTM based RNN model.

\subsection{Artificial Neural Networks (ANNs)}

ANNs \cite{mcculloch1943logical} are motivated from the human brain and consist of basic units called neurons which are connected to each other through various connections. At a neuron, incoming information is processed and passed on to connected neurons. Neurons are partitioned in various layers and generally a neuron in a given layer is connected to all neurons in the preceding and succeeding layers. The information flow between neurons is a composition of a non-linear function with a weighted linear sum of the incoming input. Generally, the non-linear function in composition is fixed at each neuron and called an activation function. The weights assigned to the connecting edges are updated in a way to minimize the suitably chosen loss function, which is done through the Gradient Descent algorithm \cite{kiefer1952} by back-propagating the gradients and updating the weights on the way. The intermediate layers between input layer and output layer containing the neurons are called hidden layers. 
ANN model has successfully able to capture non-linear complex relationship between input and output, but it is not the best choice for time-series data as the model is not designed to learn from sequential data. In what follows we will discuss RNN has the ability of information persistence.


\subsection{Recurrent Neural Networks (RNNs)}

\begin{figure}[!h]
\centering
\includegraphics[width=0.48\textwidth]{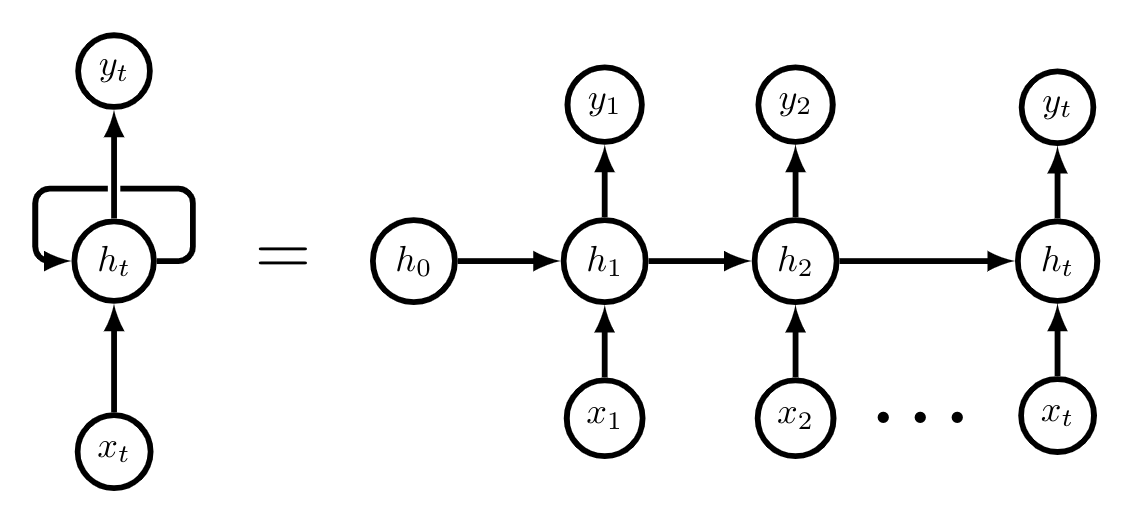} 
\caption{Structure of RNN.}
\label{rnnfig}
\end{figure}

RNN \cite{rnn1, rnn3, rnn4} are like an ANN with internal connections that enables the network to learn not just from current input but also from the previous inputs which makes it suitable for the time series data. An internal state is kept and updated over time that  stores the learning from the previous inputs and used along with current input to determine current output. A simple RNN structure is described in Figure~\ref{rnnfig}.
Theoretically, RNN can remember information through long time series data but in practice they are good in remembering information from only few steps back. RNN are prone to vanishing gradient or exploding gradient problem where the gradient decreases or increases exponentially. This problem can be avoided by using LSTM cells in RNN.

\subsection{Long Short-Term Memory (LSTM) Networks}

LSTM network \cite{lstm0, lstm1, lstm2, lstm3} overcome the shortcomings of RNN by using three inner cell gates and maintaining a memory cell to handle long term dependencies. LSTM cell can selectively read, selectively write and selectively forget. A general LSTM cell is mainly consists of four gates- an input gate to process newly coming data, a memory cell input gate to process the output of the previous LSTM cell, a forget gate to decide what to be forget and decides the optimal time lag to remember previous states, and an output gate to process all the newly calculated information and generate output. 

\subsection{Stacked LSTM Networks}
Stacked LSTM or Deep LSTM \cite{6638947} networks consist of multiple hidden layers where a layer is stacked on top of the previous layer. Each layer consists of multiple LSTM cells. A LSTM layer provides a sequence output in place of a single output to the below LSTM layer. This structure helps in better learning in sequence and time series data.

\subsection{Bi-directional LSTM Networks}
As the name suggests, a Bi-directional LSTM (BiLSTM) \cite{BiLstm} learns in both directions- forward and backward. It has two separate LSTM layers, in opposite directions of each other that helps in future to past and past to future learning. One layer takes the input in the forward direction and other in the backward direction and both layers are connected to the output layer. 


\section{Data}

Various regional centers across the world keep track of tropical cyclones and this dataset is generally known as Best Track Data (BTD). The Regional Specialised Meteorological Centre (RSMC) of IMD in New Delhi is responsible for cyclones monitoring over NIO. The yearly data from 1982 to 2020 (till June) is available on \footnote{\url{http://www.rsmcnewdelhi.imd.gov.in/index.php?option=com_content&view=article&id=48&Itemid=194&lang=en}}. The center classified a cyclonic disturbance as a tropical cyclone, when the associated MSWS is 34 knots or more \cite{mbn2013eval}. We use this BTD as dataset for our model.

\begin{figure}[!h]
\centering
\includegraphics[width=0.48\textwidth]{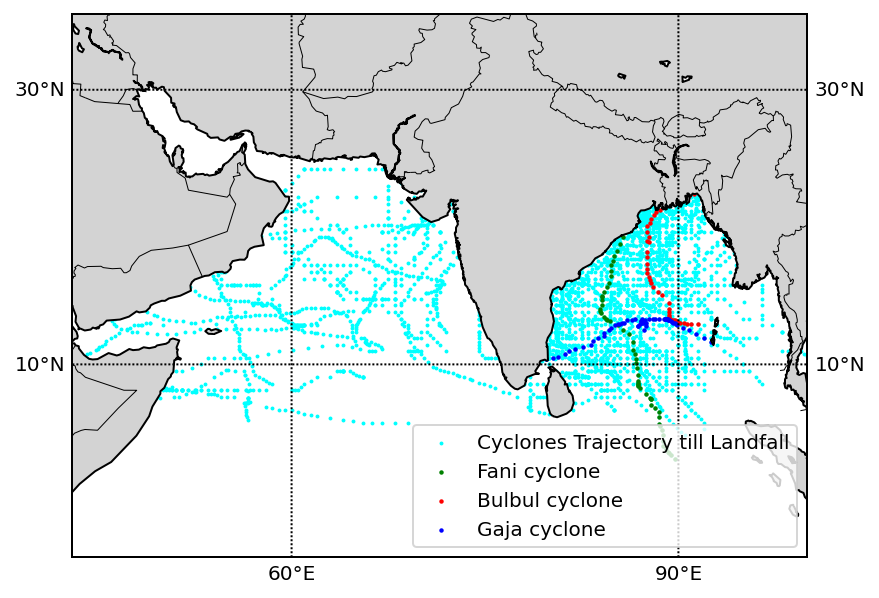} 
\caption{Cyclones trajectory till Landfall.}
\label{datafig}
\end{figure}

BTD contain many features associated with a cyclone but we only use latitude, longitude, MSWS, and estimated central pressure (ECP) as features in our model. Two more derived features distance and direction \footnote{\url{https://www.movable-type.co.uk/scripts/latlong.html}} of change between two successive recordings of a cyclone is calculated. These two features play an important role in capturing speed and direction of change of a cyclone which can be crucial for the landfall prediction. Another important factor that affects the course of a tropical cyclone is Sea Surface Temperature (SST), which is obtained from NOAA dataset provided at\footnote{\url{http://apdrc.soest.hawaii.edu/erddap/griddap/hawaii_soest_afc8_9785_907e.html}}. The above stated features are readily available during the progress of a cyclone, unlike the reanalysis data or satellite images, which required co-ordination among various agencies. 

The dataset contains few manual errors which have been corrected carefully after which a total of 6474 recordings, recorded at an interval of 3 hours, of 353 cyclones have been extracted. If the difference between two available time points for a cyclone is more than 3 hours then we have filled up missing time points to make the data a continuous time series data recorded at an interval of 3 hours. If a time series data, $d(t)$ is available for $t=t_0$ and $t_{3n}$ but missing for $t=t_3, t_6, \dots, t_{3(n-1)}$ then we evaluate $D=(d(t_{3n})-d(t_0))/n$ and fill the missing data with $d(t_{3k})=d(t_0)+kD$ for $1\leq k\leq n-1$. After completing this process and deleting any possible error in dataset we get a total number of 9088 recordings of 352 cyclones. As we are interested in predicting the landfall and therefore, for every cyclone, we retain recordings only till the landfall time. This further reduces our dataset to 3988 recordings of 206 cyclones (rest of the cyclones do not hit the coast and die out in sea). For cyclones with at least 24 hours or 8 recordings, the average time to landfall is around 80 hours. The trajectory of all cyclones till the Landfall is shown in Figure~\ref{datafig}. We do not use data of three recent devastating TCs Bulbul, Fani, and Gaja in training the model and keep them for testing our model. The trajectories of these three TCs are highlighted in  Figure~\ref{datafig}. 

\subsection{Generation of training dataset}

For a fixed cyclone, let $T_L$ be the number of data points recorded after which the landfall occurs. If we want our model to provide predictions after taking $T$ number of data points as input, we need to make sure that our model trains on inputs of size $T$. To achieve that for each cyclone, we create $T_L-T+1$ inputs. A single input is a sequence of $T$ vectors of the form
\begin{align*}
&(\operatorname{MSWS}(t), \operatorname{ECP}(t), \operatorname{SST}(t), \operatorname{distance}(t), \operatorname{direction}(t), \\ 
&\qquad \qquad  \operatorname{latitude} (t), \operatorname{longitude}(t) )
\end{align*}
where $k\leq t\leq T+k-1$. As $k$ varies from $1$ to $T_L-T+1$, we get all such inputs for a given cyclone. The target variables for each input are MSWS (in knots) at landfall, latitude and longitude at landfall, and time (in hours (h)) remaining to landfall of the cyclone to which the input corresponds to. For example, recording of data for cyclone Amphan started 00 hours on 16 May, 2020 and the landfall occurred at 12 hours on 20 May, 2020. Therefore, recordings of 108 hours are available for Amphan cyclone which amounts to $T_L=108/3= 36$. Suppose, we want to create a model for $T = 4$, then Amphan cyclone will provide $36 - 4 + 1 = 33$ data points for training the model. For a given $T$, the training dataset is collection of all such inputs across all the cyclones. Notice that each input is a time series data of length $T$.

\section{Training and Proposed Model Implementation}
\begin{figure}[!h]
\hfill
\subfigure[Model-1]{\includegraphics[width = 0.2\textwidth, height = 0.3\textheight]{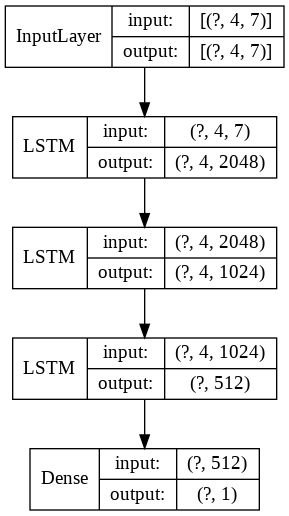}}
\hfill
\subfigure[Model-2]{\includegraphics[width = 0.25\textwidth, height = 0.3\textheight]{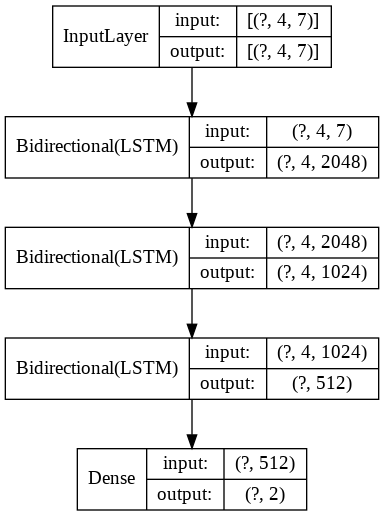}}
\hfill
\caption{RNN based on LSTM and BiLSTM for T = 4.}
\label{models}
\end{figure}

We have used two different RNN models based on LSTM, Model-1 for landfall's intensity and time prediction and Model-2 for landfall's location (latitude and longitude) prediction. Model-1 has 3 stacked LSTM layers and one dense output layer. Model-2 has 3 stacked BiLSTM layers and one dense output layer. For a faster and better training of our models, we have scaled the features of data using Standard Scaler of Scikit learn library \cite{scikit-learn}. The scaling is given by the function, $f(x) = (x - \mu)/\sigma$, where $\mu$ is the mean and $\sigma$ is the standard deviation. Model-1 scales the input variables using the Standard Scaler except the target variables, intensity and time. The Model-2 scales all input variables including target variables, latitude and longitude for training. We have implemented these models in Keras API \cite{chollet2015keras} which runs on top of low level language TensorFlow \cite{tensorflow2015-whitepaper}, developed by Google. Both  models use the default learning rate 0.01 and Adaptive moment estimation (Adam) \cite{kdj2014adam} optimizer to minimize the loss. The model train the network using MSE loss function and the accuracy is measured in terms of MAE and RMSE. The definition of these error measures are as follows:
\[
   \operatorname{MSE} = \frac{1}{n}\sum_{i=1}^{n}(y_{i} - \bar{y_{i}})^{2}, \quad \operatorname{RMSE} = \sqrt{MSE},
\]
\[
\operatorname{MAE} = \frac{1}{n}\sum_{i=1}^{n}\left | y_{i} - \bar{y_{i}} \right |,
\]
where $y_i$ is the actual value and $\bar{y_i}$ is the predicted value. We have tried various standard activation functions and selected $\operatorname{Swish}(\beta x) = x. \operatorname{Sigmoid}(\beta x) $ \cite{swish} with $\beta = 2$ for Model-1 and $\operatorname{ReLU}(x) = \max(0, x)$ \cite{relu} for  the  Model-2 to optimize the accuracy. Both models use a total of 150 epochs. The structures of Model-1 and Model-2 are shown in Figure~\ref{models} which have been generated using Keras API. 

We have used the GPU available on Google Colab to run the experiments which provides one of the GPU Nvidia K80s, T4s, P4s or P100s depending on availability. On average, Model-1 takes 80 seconds and Model-2 takes 90 seconds to complete 150 epochs.

\section{Results and Analysis}

The models (Model-1 and Model-2) take certain number of, say $T$, continuous data points of a TC, anytime during the course of the TC and predict the intensity, latitude and longitude, and time of its landfall with high accuracy. We also report the distance error in kilometers (kms) from the predicted landfall location to actual landfall location. We have reported the 5-fold validation mean accuracy of our model both in terms of RMSE and MAE for $T$ = 4, 6, 8, and 12 (that is 12h, 18h, 24h, or 36h) along with standard deviation (std). To further validate the performance of our model, we have also reported the model performance for three recent devastating cyclones Bulbul, Fani, and Gaja. These three cyclones are not the part of the training dataset. The RMSE and MAE values reported for these three cyclones are average of RMSE and MAE over a sliding window of size $T$ starting from 1st data point till the landfall.


\begin{table}[!h]
\centering
\begin{tabular}{ |p{0.5cm}|p{1.5cm}|p{0.8cm}|p{0.8cm}|p{0.8cm}|p{0.8cm}| } 
 \hline
 \multicolumn{2}{|c|}{T(Hours)} & 4(12) & 6(18) & 8(24) & 12(36) \\
 \hline
  \multicolumn{2}{|c|}{Size of dataset} & 3189 & 2843 & 2544 & 2039 \\
 \hline
 \multirow{4}{4em}{\rotatebox{90}{RMSE ($\pm$ std)}} & 5-fold Validation & 9.35 $ \qquad  \pm$ (0.63)  & 7.84  $ \qquad  \pm$ (1.09) & 7.31 $ \qquad  \pm$ (0.71) & 6.19 $ \qquad  \pm$ (0.69)\\[0.6em]
  \cline{2-6}
  & Fani &  2.66 & 1.72 & 3.43 & 5.53  \\ 
  \cline{2-6}
 & Gaja &  4.34 &  3.37 &  4.78 & 4.29  \\ 
   \cline{2-6}
 & Bulbul &  4.35 &  3.63 &  3.86 & 3.40 \\ 
 \hline
 \multirow{4}{0.6cm}{\rotatebox{90}{MAE $\pm$ (std)}} & 5-fold Validation &  5.17 $ \qquad  \pm$ (0.51) & 4.01 $ \qquad  \pm$ (0.30) & 4.24 $ \qquad  \pm$ (0.40) & 3.87 $ \qquad  \pm$ (0.36) \\[0.6em]
  \cline{2-6}
  & Fani &  2.03 &  1.37 & 2.64 &  4.10 \\ 
  \cline{2-6}
 & Gaja &  2.85 &  2.15 & 3.51 & 3.47  \\ 
   \cline{2-6}
 & Bulbul &  2.30 & 1.68 & 2.27 & 2.35 \\ 
   \hline
\end{tabular}
\caption{RMSE and MAE for landfall's intensity prediction for different values of $T$.}
\label{intensity}
\end{table}

\begin{table}[!h]
\centering
\begin{tabular}{ |p{0.5cm}|p{1.5cm}|p{0.8cm}|p{0.8cm}|p{0.8cm}|p{0.8cm}| } 
 \hline
 \multicolumn{2}{|c|}{T(Hours)} & 4(12) & 6(18) & 8(24) & 12(36) \\
 \hline
  \multicolumn{2}{|c|}{Size of dataset} & 3189 & 2843 & 2544 & 2039 \\
 \hline
 \multirow{4}{4em}{\rotatebox{90}{RMSE $\pm$ (std)}} & 5-fold Validation & 11.21 $ \qquad  \pm$ (1.03) & 10.14 $ \qquad  \pm$ (1.78) & 8.08 $ \qquad  \pm$ (0.95) & 8.52 $ \qquad  \pm$ (1.51)  \\[0.6em] 
  \cline{2-6}
  & Fani &     3.6 & 2.03 & 2.8 & 5.34 \\

  \cline{2-6}
 & Gaja &   6.0 & 5.65 & 4.0 & 6.17  \\ 
   \cline{2-6}
 & Bulbul &   4.0 & 3.4 & 3.25 & 6.21  \\ 
 \hline
 \multirow{4}{0.6cm}{\rotatebox{90}{MAE $\pm$ (std) }} & 5-fold Validation &   6.25 $ \qquad  \pm$ (0.25) & 5.26 $ \qquad  \pm$ (0.16) & 4.5 $ \qquad  \pm$ (0.58) & 5.42 $ \qquad  \pm$ (1.17) \\[0.6em]
  \cline{2-6}
  & Fani &   2.6 & 1.37 & 1.8 & 4.70  \\ 
  \cline{2-6}
 & Gaja &   3.2 & 2.74 &  3.0 & 4.90  \\ 
   \cline{2-6}
 & Bulbul &   2.3 & 1.9 & 1.77 & 5.08  \\ 
   \hline
\end{tabular}
\caption{RMSE and MAE of landfall's time prediction for different values of $T$.}
\label{time}
\end{table}

\begin{table}[!h]
\centering
\begin{tabular}{ |p{0.3cm}|p{1.45cm}|p{0.7cm}|p{0.7cm}|p{0.7cm}|p{0.7cm}|p{0.8cm}| } 
 \hline
 \multicolumn{3}{|c|}{T(Hours)} & 4(12) & 6(18) & 8(24) & 12(36) \\
 \hline
  \multicolumn{3}{|c|}{Size of dataset} & 3189 & 2843 & 2544 & 2039 \\
 \hline
 \multirow{8}{4em}{\rotatebox{90}{RMSE $\pm$ (std) }} & \multirow{2}{4em}{5-fold Validation} & Lati & 0.95 $ \qquad  \pm$ (0.04) & 0.67 $ \qquad  \pm$ (0.06) & 0.52 $ \qquad  \pm$ (0.12) &0.39  $ \qquad  \pm$ (0.15)\\[0.6em]
  \cline{3-7}
  & & Long &  1.30 $ \qquad  \pm$ (0.14) & 0.85 $ \qquad  \pm$ (0.04)& 0.72 $ \qquad  \pm$ (0.08) & 0.50 $ \qquad  \pm$ (0.10)\\[0.6em]
  \cline{2-7}
  & \multirow{2}{4em}{Fani}&  Lati  &  0.33 & 0.16 & 0.11 & 0.13 \\
   \cline{3-7}
   & & Long &  0.60 & 0.36 & 0.19 & 0.23 \\
  \cline{2-7}
    & \multirow{2}{4em}{Gaja}&  Lati  &  0.85 & 0.53 & 0.22 & 0.13 \\
   \cline{3-7}
   & & Long &  0.45 & 0.22 & 0.07 & 0.09  \\
  \cline{2-7}
    & \multirow{2}{4em}{Bulbul}&  Lati  &  0.19 & 0.15 & 0.10 & 0.09 \\
   \cline{3-7}
   & & Long &  0.37 & 0.26 & 0.17 & 0.19 \\
  \cline{2-7}
 \hline
 \multirow{8}{4em}{\rotatebox{90}{MAE $\pm$ (std)}} & \multirow{2}{4em}{5-fold Validation} & Lati & 0.52  $ \qquad  \pm$ (0.02) & 0.33 $ \qquad  \pm$ (0.01) & 0.24 $ \qquad  \pm$ (0.02)& 0.19 $ \qquad  \pm$ (0.01)\\[0.6em]
  \cline{3-7}
  & & Long &  0.75 $ \qquad  \pm$ (0.05) & 0.46 $ \qquad  \pm$ (0.02) & 0.37 $ \qquad  \pm$ (0.02) & 0.29 $ \qquad  \pm$ (0.02)\\[0.6em]
  \cline{2-7}
  & \multirow{2}{4em}{Fani}&  Lati  & 0.27 & 0.11 & 0.08 & 0.10 \\
   \cline{3-7}
   & & Long & 0.41 & 0.26 & 0.14 & 0.19 \\
  \cline{2-7}
    & \multirow{2}{4em}{Gaja}&  Lati  &  0.28 & 0.16 & 0.10 & 0.1 \\
   \cline{3-7}
   & & Long &  0.15 & 0.09 & 0.05 & 0.07 \\
  \cline{2-7}
    & \multirow{2}{4em}{Bulbul}&  Lati  &  0.15 & 0.09 & 0.07 & 0.07 \\
   \cline{3-7}
   & & Long &  0.29 & 0.19 & 0.14 & 0.16
 \\[1em]
 \hline
  \multirow{4}{5em}{\rotatebox{90}{Distance $\pm$ (std)}} & \multicolumn{2}{|c|}{5-fold Validation} & 106.3 $ \qquad  \pm$ (5.79) & 67.0 $ \qquad  \pm$ (2.51) & 51.7 $ \qquad  \pm$ (1.20)& 41.2 $ \qquad  \pm$ (3.12)\\[0.6em]
  \cline{2-7}
  & \multicolumn{2}{|c|}{ Fani}   &  56.1 & 32.4 & 18.7 & 24.6 \\
  \cline{2-7}
    & \multicolumn{2}{|c|}{Gaja}   & 38.5 & 22.7 & 15.1 & 15.1
 \\
  \cline{2-7}
    & \multicolumn{2}{|c|}{Bulbul} & 37.9 & 24.7 & 18.2 & 19.9 \\
   \cline{3-7}
 \hline
\end{tabular}
\caption{RMSE, MAE and Distance Error (kilometers) for Landfall Location Prediction for different T's}
\label{location}
\end{table}

\begin{figure*}[!h]
\centering
\includegraphics[width=0.78\textwidth]{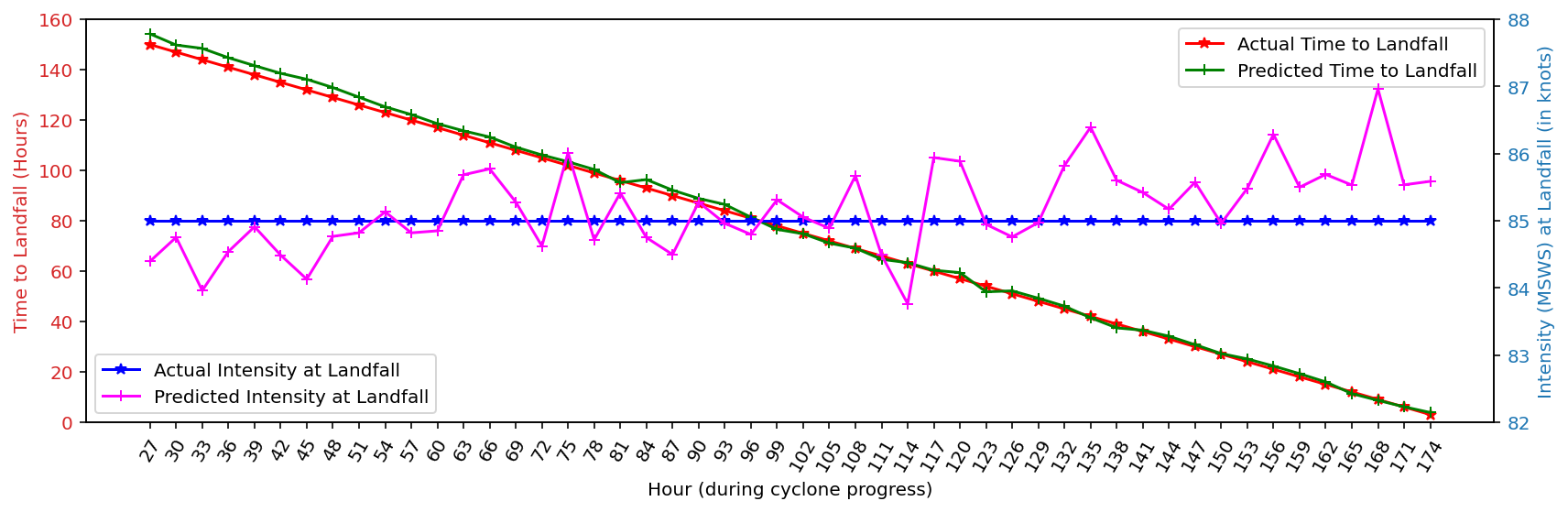} 
\caption{Predicted and actual intensity and time of landfall of Fani for $T = 8$.}
\label{fanifig}
\end{figure*}

\begin{figure*}[!h]
\centering
\includegraphics[width=0.78\textwidth]{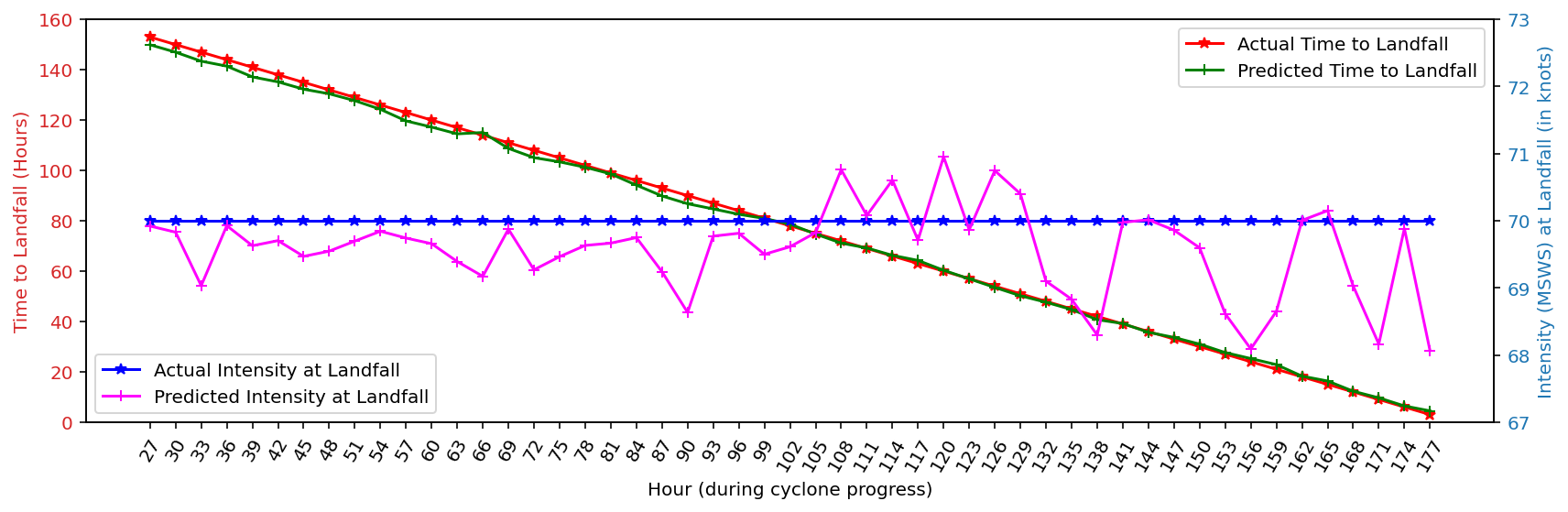} 
\caption{Predicted and actual intensity and time of landfall of Gaja for $T = 8$.}
\label{gajafig}
\end{figure*}

\begin{figure*}[!h]
\centering
\includegraphics[width=0.78\textwidth]{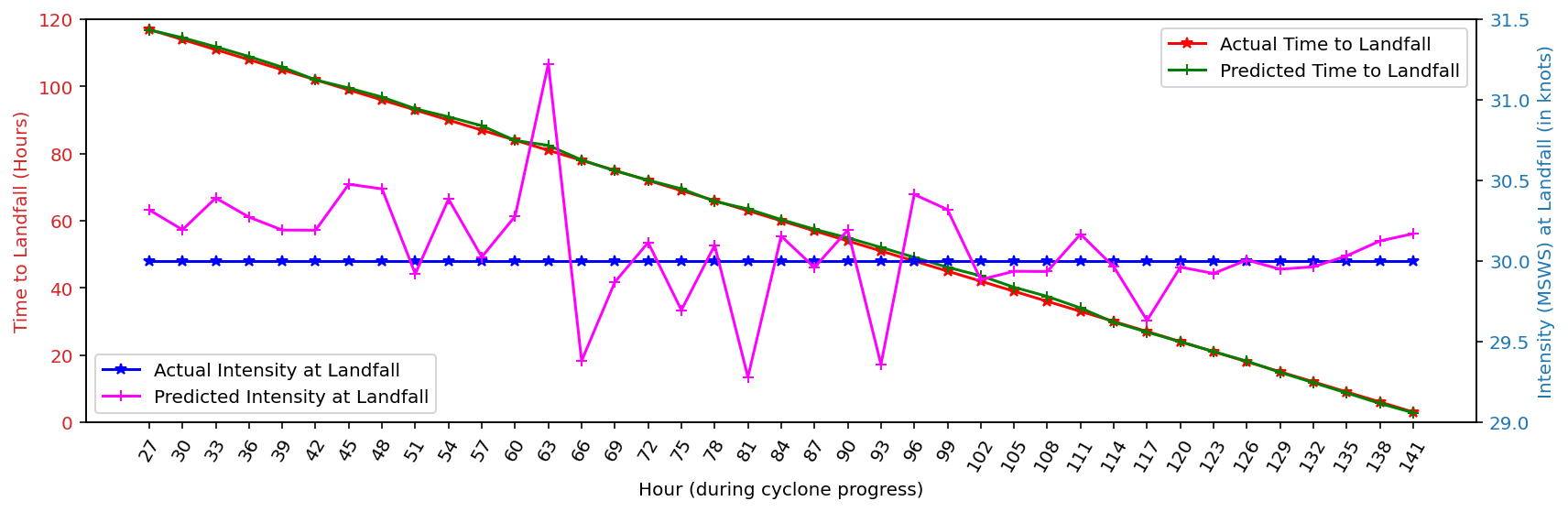} 
\caption{Predicted and actual intensity and time of landfall of Bulbul for $T = 8$.}
\label{bulbulfig}
\end{figure*}

\begin{table}[!h]
\centering
\begin{tabular}{ |p{3.4cm}|p{0.7cm}|p{0.7cm}|p{0.7cm}|p{0.7cm}| } 
  \hline
  Lead Time (hours) & 36 & 48 & 60 & 72 \\
 \hline
  Landfall Time MAE &  4.96 &  5.53 & 6.8 &  9.6 \\ 
   \hline
 Landfall Distance MAE &  42.84 &  78.08 & 92.6 & 112.5  \\ 
   \hline
\end{tabular}
\caption{4/5 year (2015-2019) average accuracy reported by IMD for cyclones in NIO.}
\label{imdData}
\end{table}

In Tables~\ref{intensity} and \ref{time}, the RMSE and MAE of prediction of intensity at landfall and time remaining to landfall are reported for different values of $T$, respectively, along with the size of training dataset. For $T = 8$, that is if 24 hours data of cyclone is used then the intensity and time can be predicted within an MAE of 4.24 knots and 4.5 hours, respectively. From Table~\ref{grade}, we can see that the range of MSWS for all Grades, except Grade 2, is at least 10, this implies that with a very high probability, the model will predict the correct intensity grade at landfall of a TC. Moreover, since we obtain such good accuracy with only 24 hours of observation and the landfall occurs on average at the 80th hour in NIO, the model can help authorities to prepare well advance in time to take any action. The performance of model is even better than 5-fold validation accuracy for cyclones Bulbul, Fani, and Gaja as evident from Tables~\ref{intensity} and \ref{time}.

In Figures \ref{fanifig}, \ref{gajafig}, and \ref{bulbulfig}, the predicted intensity and actual intensity along with predicted time to landfall and actual time to Landfall are shown for $T = 8$ (24 hours) for cyclones Bulbul, Fani, and Gaja, respectively. To obtain these figures, we choose a sliding window of 24 hours and get the prediction from the model. For example, the values at 27th hour and 75th hour are the predictions using the data between 0th and 24th hours and 48th and 72nd hours, respectively. It is evident that the model has consistently performed well irrespective of whether the prediction point is close to the landfall or far from the landfall. One should note that these three cyclones took a long time ($>$141 hours) to hit the coastal region, despite this the model's predictions are consistently good even at the beginning of the cyclone.

In Table~\ref{location}, the RMSE and MAE of latitude and longitude prediction (in degrees) at landfall are reported for different values of $T$. A slight error in latitude and longitude may lead to an error of several kilometers in the location. Therefore, we also report the corresponding distance error in kilometers. The distance error is calculated using the distance between actual and predicted landfall location. For example, for $T = 8$, the model can predict the landfall location with an error of 51.7 kms. 

\begin{figure*}[!h]
\centering
\includegraphics[width=0.78\textwidth]{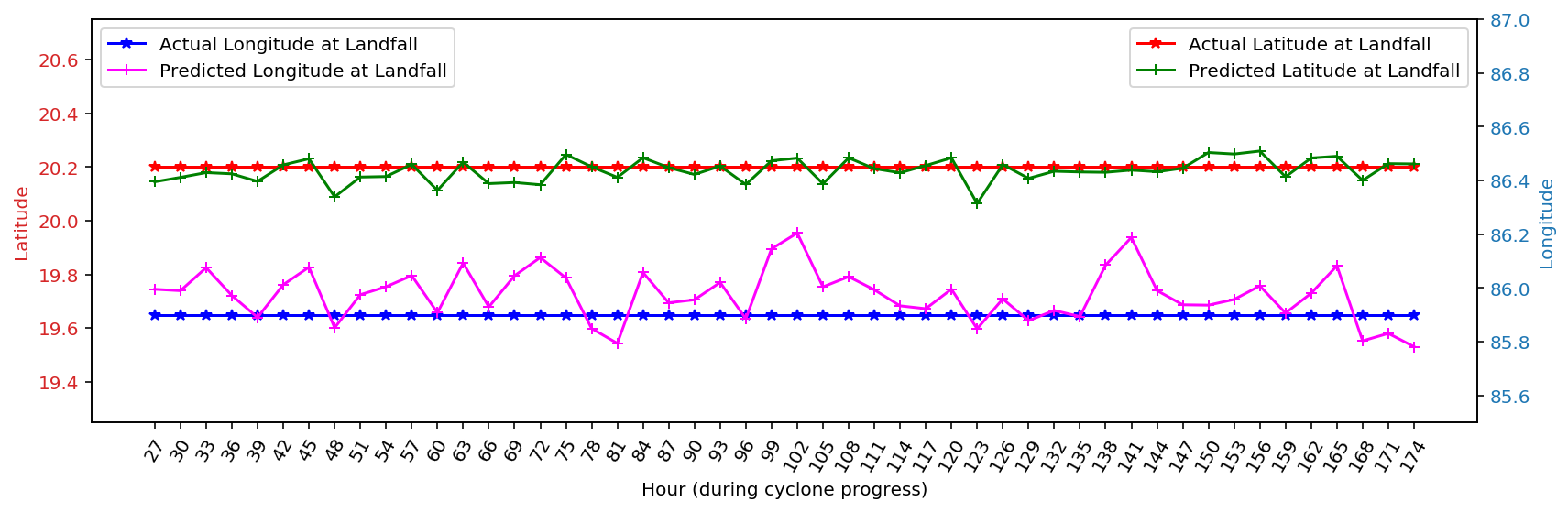} 
\caption{Predicted and Actual Landfall Latitude/Longitude of Fani cyclone for T = 8}
\label{fanifig1}
\end{figure*}

\begin{figure*}[!h]
\centering
\includegraphics[width=0.78\textwidth]{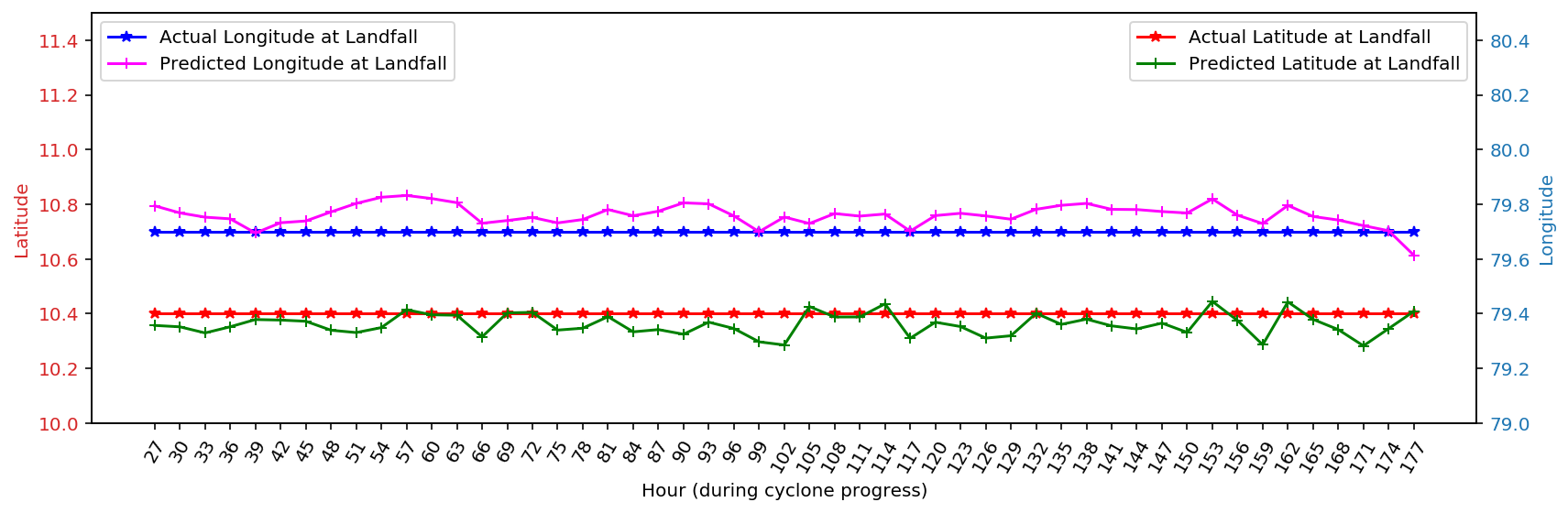} 
\caption{Predicted and Actual Landfall Latitude/Longitude of Gaja cyclone for T = 8}
\label{gajafig1}
\end{figure*}

\begin{figure*}[!h]
\centering
\includegraphics[width=0.78\textwidth]{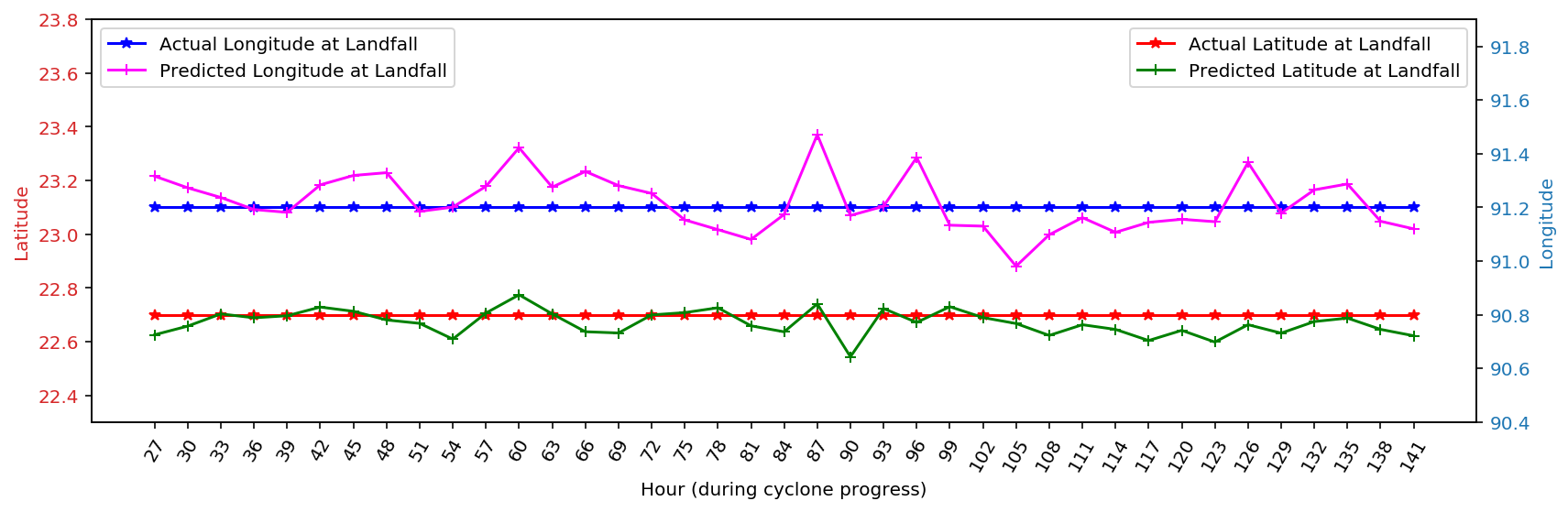} 
\caption{Predicted and Actual Landfall Latitude/Longitude of Bulbul cyclone for T = 8}
\label{bulbulfig1}
\end{figure*}

In the Figures \ref{fanifig1}, \ref{gajafig1}, \ref{bulbulfig1}, the predicted latitude, longitude, and actual latitude, longitude at landfall are shown for $T = 8$ for cyclones Bulbul, Fani, and Gaja, respectively. It is once again evident that the model has consistently performed well irrespective of whether the prediction point is close to landfall or far from landfall. 

The standard deviation among 5-fold for all the three prediction problems is quite low, means the model performs consistently and it can be reliably deployed for above stated prediction problems. We compare our results with the landfall forecasts reported by IMD on its website\footnote{\url{http://www.rsmcnewdelhi.imd.gov.in/index.php?option=com_content&view=article&id=45&Itemid=191&lang=en}}. IMD provides average forecasting error of location and time of landfall for each year starting from 2003. Error values reported from earlier years are large and are not suitable for a fair comparison. Therefore, we have calculated last 4 or 5 years (as per data availability) average MAE scores reported at the IMD website and provided in the Table~\ref{imdData}. The lead time of X hours in Table~\ref{imdData} means that forecast was done X hours before the landfall. MAE reported for our model in Tables~\ref{location} and \ref{time} are either less or comparable to values in Table~\ref{imdData}. We would like to emphasize that our testing set may include any cyclone starting from 1982 and this may further would have increased the 5-fold validation error values. Another likely reason behind the higher 5-fold validation error is the re-curving cyclones. Generally, the TC's related studies do not consider cyclones with loops. Our model do not have any such restriction. Further, when we compare the results for recent cyclones like Bulbul (2019), Fani (2019), and Gaja (2018), the difference between errors reported by us and IMD is striking. For example, for Bulbul, even for a large lead time, say 72 hours, we can predict time and location of its landfall within an error of 1.77 hours and 18.2 kms, respectively while the corresponding errors from IMD are 9.6 hours and 112.5 kms, see Table~\ref{imdData}.

\begin{table}[!h]
\centering
\begin{tabular}{ |p{3cm}|p{0.8cm}|p{1.9cm}|p{1.3cm}| } 
  \hline
  Target/Model & ANN & RNN (GRU) & 1D-CNN \\
 \hline
  Intensity (knots) &  10.78 &  4.8 & 9.33  \\ 
   \hline
  Landfall Time (hrs) &  9.55 &  5.21 & 9.66 \\ 
   \hline
    Lati (degree)&  0.57 &  0.38 & 0.64  \\ 
   \hline
       Long (degree) &  0.84 &  0.45 & 0.96  \\ 
   \hline
       Distance(kms) &  118.4 &  61.8 & 135.04 \\ 
   \hline
\end{tabular}
\caption{5-fold Validation MAE of ANN, RNN(GRU based) and 1D-CNN models for T = 8.}
\label{modlesPerform}
\end{table}

The performance of few other models in mentioned in Table~\ref{modlesPerform}. The ANN model has 5 hidden layers of sizes 1024, 512, 256, 128 and 32 with activation function $\operatorname{Swish(2x)}$. The GRU based RNN model has same configurations as the earlier defined LSTM models. The 1D-CNN \cite{KY20141DCNN} model has 2 convolutional layers each with 512 filters, with Batch-Normalization \cite{pmlr-v37-ioffe15}, Dropout \cite{SNGAS2014Dropout}, a max pool layer, 8 dense layers each of size 512, with $\operatorname{ReLU}$  activation function and Adam optimizer.

\section{Conclusion}

We presented a model which used LSTM network based on RNN to predict the intensity, location, and time of landfall of a tropical cyclone in the North Indian Ocean. The model predicts the landfall characteristics with high accuracy and beat the model used by India Meteorological Department on recent cyclones. The biggest advantage of this model over earlier models is that it is trained to provide prediction using only few continuous data points taken from anywhere from the course of the cyclone. Using our model, the landfall characteristics of a cyclone can be predicted with high accuracy after a few hours after the origin of the cyclone which will provide ample time to disaster managers to decide if they need to evacuate certain areas (depending on intensity prediction) and if they need to then precisely when (depending on time prediction) and which area (depending on location prediction). In a future work, we want to include the characteristics of terrain from which the cyclone is passing as an input feature in the model. We also want to extend this work for Atlantic and Pacific oceans. 

\bibliography{reference.bib}

\begin{thebibliography}{35}
\providecommand{\natexlab}[1]{#1}
\providecommand{\url}[1]{\texttt{#1}}
\providecommand{\urlprefix}{URL }
\expandafter\ifx\csname urlstyle\endcsname\relax
  \providecommand{\doi}[1]{doi:\discretionary{}{}{}#1}\else
  \providecommand{\doi}{doi:\discretionary{}{}{}\begingroup
  \urlstyle{rm}\Url}\fi

\bibitem[{Abadi et~al.(2015)Abadi, Agarwal, Barham, Brevdo, Chen, Citro,
  Corrado, Davis, Dean, Devin, Ghemawat, Goodfellow, Harp, Irving, Isard, Jia,
  Jozefowicz, Kaiser, Kudlur, Levenberg, Man\'{e}, Monga, Moore, Murray, Olah,
  Schuster, Shlens, Steiner, Sutskever, Talwar, Tucker, Vanhoucke, Vasudevan,
  Vi\'{e}gas, Vinyals, Warden, Wattenberg, Wicke, Yu, and
  Zheng}]{tensorflow2015-whitepaper}
Abadi, M.; Agarwal, A.; Barham, P.; Brevdo, E.; Chen, Z.; Citro, C.; Corrado,
  G.~S.; Davis, A.; Dean, J.; Devin, M.; Ghemawat, S.; Goodfellow, I.; Harp,
  A.; Irving, G.; Isard, M.; Jia, Y.; Jozefowicz, R.; Kaiser, L.; Kudlur, M.;
  Levenberg, J.; Man\'{e}, D.; Monga, R.; Moore, S.; Murray, D.; Olah, C.;
  Schuster, M.; Shlens, J.; Steiner, B.; Sutskever, I.; Talwar, K.; Tucker, P.;
  Vanhoucke, V.; Vasudevan, V.; Vi\'{e}gas, F.; Vinyals, O.; Warden, P.;
  Wattenberg, M.; Wicke, M.; Yu, Y.; and Zheng, X. 2015.
\newblock {TensorFlow}: Large-Scale Machine Learning on Heterogeneous Systems.
\newblock \urlprefix\url{https://www.tensorflow.org/}.
\newblock Software available from tensorflow.org.

\bibitem[{Alemany et~al.(2018)Alemany, Beltran, Perez, and
  Ganzfried}]{abjg2018predicting}
Alemany, S.; Beltran, J.; Perez, A.; and Ganzfried, S. 2018.
\newblock Predicting Hurricane Trajectories Using a Recurrent Neural Network.
\newblock \emph{Proceedings of the AAAI Conference on Artificial Intelligence}
  33.
\newblock \doi{10.1609/aaai.v33i01.3301468}.

\bibitem[{Chaudhuri et~al.(2017)Chaudhuri, Basu, Das, Goswami, and
  Varshney}]{chaudhuri2017swarm}
Chaudhuri, S.; Basu, D.; Das, D.; Goswami, S.; and Varshney, S. 2017.
\newblock Swarm intelligence and neural nets in forecasting the maximum
  sustained wind speed along the track of tropical cyclones over Bay of Bengal.
\newblock \emph{Natural Hazards} 87(3): 1413--1433.

\bibitem[{Chaudhuri et~al.(2015)Chaudhuri, Dutta, Goswami, and
  Middey}]{chaudhuri2015track}
Chaudhuri, S.; Dutta, D.; Goswami, S.; and Middey, A. 2015.
\newblock Track and intensity forecast of tropical cyclones over the North
  Indian Ocean with multilayer feed forward neural nets.
\newblock \emph{Meteorological Applications} 22(3): 563--575.

\bibitem[{Chollet(2015)}]{chollet2015keras}
Chollet, F. 2015.
\newblock Keras.
\newblock \url{https://github.com/fchollet/keras}.

\bibitem[{Cleeremans, Servan-Schreiber, and McClelland(1989)}]{rnn3}
Cleeremans, A.; Servan-Schreiber, D.; and McClelland, J.~L. 1989.
\newblock Finite State Automata and Simple Recurrent Networks.
\newblock \emph{Neural Comput.} 1(3): 372–381.
\newblock ISSN 0899-7667.
\newblock \doi{10.1162/neco.1989.1.3.372}.
\newblock \urlprefix\url{https://doi.org/10.1162/neco.1989.1.3.372}.

\bibitem[{Fritz et~al.(2009)Fritz, Blount, Thwin, Thu, and
  Chan}]{cyclonefhfc2009}
Fritz, H.; Blount, C.; Thwin, S.; Thu, M.; and Chan, N. 2009.
\newblock Cyclone Nargis storm surge in Myanmar.
\newblock \emph{Nature Geoscience} 2: 448--449.
\newblock \doi{10.1038/ngeo558}.

\bibitem[{{Gers} and {Schmidhuber}(2001)}]{lstm3}
{Gers}, F.~A.; and {Schmidhuber}, E. 2001.
\newblock LSTM recurrent networks learn simple context-free and
  context-sensitive languages.
\newblock \emph{IEEE Transactions on Neural Networks} 12(6): 1333--1340.

\bibitem[{{Gers}, {Schmidhuber}, and {Cummins}(1999)}]{lstm1}
{Gers}, F.~A.; {Schmidhuber}, J.; and {Cummins}, F. 1999.
\newblock Learning to forget: continual prediction with LSTM.
\newblock In \emph{1999 Ninth International Conference on Artificial Neural
  Networks ICANN 99. (Conf. Publ. No. 470)}, volume~2, 850--855 vol.2.

\bibitem[{Gers, Schraudolph, and Schmidhuber(2003)}]{lstm2}
Gers, F.~A.; Schraudolph, N.~N.; and Schmidhuber, J. 2003.
\newblock Learning Precise Timing with Lstm Recurrent Networks.
\newblock \emph{J. Mach. Learn. Res.} 3(null): 115–143.
\newblock ISSN 1532-4435.
\newblock \doi{10.1162/153244303768966139}.
\newblock \urlprefix\url{https://doi.org/10.1162/153244303768966139}.

\bibitem[{Giffard-Roisin et~al.(2018)Giffard-Roisin, Yang, Charpiat, K{\'e}gl,
  and Monteleoni}]{giffard2018deep}
Giffard-Roisin, S.; Yang, M.; Charpiat, G.; K{\'e}gl, B.; and Monteleoni, C.
  2018.
\newblock Deep learning for hurricane track forecasting from aligned
  spatio-temporal climate datasets .

\bibitem[{Giffard-Roisin et~al.(2020)Giffard-Roisin, Yang, Charpiat,
  Kumler~Bonfanti, Kégl, and Monteleoni}]{gsmgmc2020learning}
Giffard-Roisin, S.; Yang, M.; Charpiat, G.; Kumler~Bonfanti, C.; Kégl, B.; and
  Monteleoni, C. 2020.
\newblock Tropical Cyclone Track Forecasting Using Fused Deep Learning From
  Aligned Reanalysis Data.
\newblock \emph{Frontiers in Big Data} 3: 1.
\newblock ISSN 2624-909X.
\newblock \doi{10.3389/fdata.2020.00001}.
\newblock
  \urlprefix\url{https://www.frontiersin.org/article/10.3389/fdata.2020.00001}.

\bibitem[{{Graves}, {Mohamed}, and {Hinton}(2013)}]{6638947}
{Graves}, A.; {Mohamed}, A.; and {Hinton}, G. 2013.
\newblock Speech recognition with deep recurrent neural networks.
\newblock In \emph{2013 IEEE International Conference on Acoustics, Speech and
  Signal Processing}, 6645--6649.

\bibitem[{Hall and Jewson(2007)}]{htjs2007statistical}
Hall, T.; and Jewson, S. 2007.
\newblock Statistical modelling of North Atlantic tropical cyclone tracks.
\newblock \emph{Tellus A} \doi{10.3402/tellusa.v59i4.15017}.

\bibitem[{Hochreiter and Schmidhuber(1997)}]{lstm0}
Hochreiter, S.; and Schmidhuber, J. 1997.
\newblock Long Short-Term Memory.
\newblock \emph{Neural Comput.} 9(8): 1735–1780.
\newblock ISSN 0899-7667.
\newblock \doi{10.1162/neco.1997.9.8.1735}.
\newblock \urlprefix\url{https://doi.org/10.1162/neco.1997.9.8.1735}.

\bibitem[{Ioffe and Szegedy(2015)}]{pmlr-v37-ioffe15}
Ioffe, S.; and Szegedy, C. 2015.
\newblock Batch Normalization: Accelerating Deep Network Training by Reducing
  Internal Covariate Shift.
\newblock In Bach, F.; and Blei, D., eds., \emph{Proceedings of the 32nd
  International Conference on Machine Learning}, volume~37 of \emph{Proceedings
  of Machine Learning Research}, 448--456. Lille, France: PMLR.
\newblock \urlprefix\url{http://proceedings.mlr.press/v37/ioffe15.html}.

\bibitem[{Jordan(1990)}]{rnn1}
Jordan, M.~I. 1990.
\newblock \emph{Attractor Dynamics and Parallelism in a Connectionist
  Sequential Machine}, 112–127.
\newblock IEEE Press.
\newblock ISBN 0818620153.

\bibitem[{Kiefer and Wolfowitz(1952)}]{kiefer1952}
Kiefer, J.; and Wolfowitz, J. 1952.
\newblock Stochastic Estimation of the Maximum of a Regression Function.
\newblock \emph{Ann. Math. Statist.} 23(3): 462--466.
\newblock \doi{10.1214/aoms/1177729392}.
\newblock \urlprefix\url{https://doi.org/10.1214/aoms/1177729392}.

\bibitem[{Kim(2014)}]{KY20141DCNN}
Kim, Y. 2014.
\newblock Convolutional Neural Networks for Sentence Classification.
\newblock In \emph{Proceedings of the 2014 Conference on Empirical Methods in
  Natural Language Processing ({EMNLP})}, 1746--1751. Doha, Qatar: Association
  for Computational Linguistics.
\newblock \doi{10.3115/v1/D14-1181}.
\newblock \urlprefix\url{https://www.aclweb.org/anthology/D14-1181}.

\bibitem[{Kingma and Ba(2014)}]{kdj2014adam}
Kingma, D.; and Ba, J. 2014.
\newblock Adam: A Method for Stochastic Optimization.
\newblock \emph{International Conference on Learning Representations} .

\bibitem[{Krishnamurti et~al.(1999)Krishnamurti, Kishtawal, LaRow, Bachiochi,
  Zhang, Williford, Gadgil, and Surendran}]{kkcdsss1999improved}
Krishnamurti, T.; Kishtawal, C.; LaRow, T.; Bachiochi, D.; Zhang, Z.;
  Williford, C.; Gadgil, S.; and Surendran, S. 1999.
\newblock Improved Weather and Seasonal Climate Forecasts From Multi-Model
  Superensemble.
\newblock \emph{Science} 285: 1548--1550.
\newblock \doi{10.1126/science.285.5433.1548}.

\bibitem[{Kumar, Lal, and Kumar(2020)}]{turbulanceksp2020}
Kumar, S.; Lal, P.; and Kumar, A. 2020.
\newblock Turbulence of tropical cyclone ‘Fani’ in the Bay of Bengal and
  Indian subcontinent.
\newblock \emph{Natural Hazards} 103: 1613--1622.
\newblock \doi{10.1007/s11069-020-04033-5}.

\bibitem[{Leroux et~al.(2018)Leroux, Wood, Elsberry, Cayanan, Hendricks, Kucas,
  Otto, Rogers, Sampson, and Yu}]{LEROUX201885}
Leroux, M.-D.; Wood, K.; Elsberry, R.~L.; Cayanan, E.~O.; Hendricks, E.; Kucas,
  M.; Otto, P.; Rogers, R.; Sampson, B.; and Yu, Z. 2018.
\newblock Recent Advances in Research and Forecasting of Tropical Cyclone
  Track, Intensity, and Structure at Landfall.
\newblock \emph{Tropical Cyclone Research and Review} 7(2): 85 -- 105.
\newblock ISSN 2225-6032.
\newblock \doi{https://doi.org/10.6057/2018TCRR02.02}.
\newblock
  \urlprefix\url{http://www.sciencedirect.com/science/article/pii/S2225603219300189}.

\bibitem[{Maskey et~al.(2020)Maskey, Ramachandran, Ramasubramanian, Gurung,
  Freitag, Kaulfus, Bollinger, Cecil, and Miller}]{maskey2020deepti}
Maskey, M.; Ramachandran, R.; Ramasubramanian, M.; Gurung, I.; Freitag, B.;
  Kaulfus, A.; Bollinger, D.; Cecil, D.~J.; and Miller, J. 2020.
\newblock Deepti: Deep-Learning-Based Tropical Cyclone Intensity Estimation
  System.
\newblock \emph{IEEE Journal of Selected Topics in Applied Earth Observations
  and Remote Sensing} 13: 4271--4281.

\bibitem[{McCulloch and Pitts(1943)}]{mcculloch1943logical}
McCulloch, W.~S.; and Pitts, W. 1943.
\newblock A logical calculus of the ideas immanent in nervous activity.
\newblock \emph{The bulletin of mathematical biophysics} 5(4): 115--133.

\bibitem[{Mohapatra, Bandyopadhyay, and Nayak(2013)}]{mbn2013eval}
Mohapatra, M.; Bandyopadhyay, B.; and Nayak, D. 2013.
\newblock Evaluation of operational tropical cyclone intensity forecasts over
  north Indian Ocean issued by India Meteorological Department.
\newblock \emph{Natural Hazards} 68.
\newblock \doi{10.1007/s11069-013-0624-z}.

\bibitem[{Moradi~Kordmahalleh, Gorji~Sefidmazgi, and
  Homaifar(2016)}]{moradi2016sparse}
Moradi~Kordmahalleh, M.; Gorji~Sefidmazgi, M.; and Homaifar, A. 2016.
\newblock A sparse recurrent neural network for trajectory prediction of
  atlantic hurricanes.
\newblock In \emph{Proceedings of the Genetic and Evolutionary Computation
  Conference 2016}, 957--964.

\bibitem[{Nair and Hinton(2010)}]{relu}
Nair, V.; and Hinton, G.~E. 2010.
\newblock Rectified Linear Units Improve Restricted Boltzmann Machines.
\newblock In F{\"{u}}rnkranz, J.; and Joachims, T., eds., \emph{Proceedings of
  the 27th International Conference on Machine Learning (ICML-10), June 21-24,
  2010, Haifa, Israel}, 807--814. Omnipress.
\newblock \urlprefix\url{https://icml.cc/Conferences/2010/papers/432.pdf}.

\bibitem[{{Pearlmutter}(1989)}]{rnn4}
{Pearlmutter}. 1989.
\newblock Learning state space trajectories in recurrent neural networks.
\newblock In \emph{International 1989 Joint Conference on Neural Networks},
  365--372 vol.2.

\bibitem[{Pedregosa et~al.(2011)Pedregosa, Varoquaux, Gramfort, Michel,
  Thirion, Grisel, Blondel, Prettenhofer, Weiss, Dubourg, Vanderplas, Passos,
  Cournapeau, Brucher, Perrot, and Duchesnay}]{scikit-learn}
Pedregosa, F.; Varoquaux, G.; Gramfort, A.; Michel, V.; Thirion, B.; Grisel,
  O.; Blondel, M.; Prettenhofer, P.; Weiss, R.; Dubourg, V.; Vanderplas, J.;
  Passos, A.; Cournapeau, D.; Brucher, M.; Perrot, M.; and Duchesnay, E. 2011.
\newblock Scikit-learn: Machine Learning in {P}ython.
\newblock \emph{Journal of Machine Learning Research} 12: 2825--2830.

\bibitem[{{Pradhan} et~al.(2018){Pradhan}, {Aygun}, {Maskey}, {Ramachandran},
  and {Cecil}}]{RRMRD2018}
{Pradhan}, R.; {Aygun}, R.~S.; {Maskey}, M.; {Ramachandran}, R.; and {Cecil},
  D.~J. 2018.
\newblock Tropical Cyclone Intensity Estimation Using a Deep Convolutional
  Neural Network.
\newblock \emph{IEEE Transactions on Image Processing} 27(2): 692--702.
\newblock \doi{10.1109/TIP.2017.2766358}.

\bibitem[{Ramachandran, Zoph, and Le(2017)}]{swish}
Ramachandran, P.; Zoph, B.; and Le, Q.~V. 2017.
\newblock Searching for Activation Functions.

\bibitem[{{Schuster} and {Paliwal}(1997)}]{BiLstm}
{Schuster}, M.; and {Paliwal}, K.~K. 1997.
\newblock Bidirectional recurrent neural networks.
\newblock \emph{IEEE Transactions on Signal Processing} 45(11): 2673--2681.

\bibitem[{Srivastava et~al.(2014)Srivastava, Hinton, Krizhevsky, Sutskever, and
  Salakhutdinov}]{SNGAS2014Dropout}
Srivastava, N.; Hinton, G.; Krizhevsky, A.; Sutskever, I.; and Salakhutdinov,
  R. 2014.
\newblock Dropout: A Simple Way to Prevent Neural Networks from Overfitting
  15(1): 1929–1958.
\newblock ISSN 1532-4435.

\bibitem[{Wang et~al.(2009)Wang, Ha, Kumar, Wang, Long, Chelliah, Bell, and
  Peng}]{whaw2009statistical}
Wang, H.; Ha, K.-J.; Kumar, A.; Wang, W.; Long, L.; Chelliah, M.; Bell, G.; and
  Peng, P. 2009.
\newblock A Statistical Forecast Model for Atlantic Seasonal Hurricane Activity
  Based on the NCEP Dynamical Seasonal Forecast.
\newblock \emph{Journal of Climate - J CLIMATE} 22: 4481--4500.
\newblock \doi{10.1175/2009JCLI2753.1}.

\end{thebibliography}

\end{document}